\documentclass[11pt,a4paper]{article}
\usepackage{times,latexsym}
\usepackage{url}
\usepackage[T1]{fontenc}
\usepackage{caption}
\usepackage{subcaption}
\usepackage{multirow}
\usepackage{tikz}
\usepackage{xcolor}
\usetikzlibrary{shapes.geometric, arrows}
\usetikzlibrary{arrows.meta}

\tikzstyle{arrow} = [thick,->,>=stealth]

\usepackage[utf8]{inputenc}

\usepackage{microtype}

\usepackage{graphicx}

\usepackage{comment}

\usepackage{tacl2021v1}

\usepackage{xspace,mfirstuc,tabulary}

\newif\iftaclinstructions
\taclpubformattrue 
\iftaclinstructions
\renewcommand{\confidential}{}
\renewcommand{\anonsubtext}{(No author info supplied here, for consistency with
TACL-submission anonymization requirements)}
\newcommand{\instr}
\fi

\iftaclpubformat 

\else

\fi

\title{Don’t Learn, Ground: A Case for Natural Language Inference with Visual Grounding}

\author{
  Daniil Ignatev$^\diamond$$^\dagger$ 
  \and
  Ayman Santeer$^\diamond$
  \and
  Albert Gatt$^\diamond$
  \and
  Denis Paperno$^\diamond$
  \\
  \ \\
  $^\diamond$Utrecht University
  \\
  \ \\
  \\
  $^\dagger$Corresponding author: \texttt{d.ignatev@uu.nl}
}

\date{}

\begin{document}
\maketitle
\begin{abstract}
We propose a zero-shot method for Natural Language Inference (NLI) that leverages multimodal representations by grounding language in visual contexts. Our approach generates visual representations of premises using text-to-image models and performs inference by comparing these representations with textual hypotheses. We evaluate two inference techniques: cosine similarity and visual question answering. Our method achieves high accuracy without task-specific fine-tuning, demonstrating robustness against textual biases and surface heuristics. Additionally, we design a controlled adversarial dataset to validate the robustness of our approach. Our findings suggest that leveraging visual modality as a meaning representation provides a promising direction for robust natural language understanding. 
\end{abstract}

\begin{figure*}
    \centering
    \input{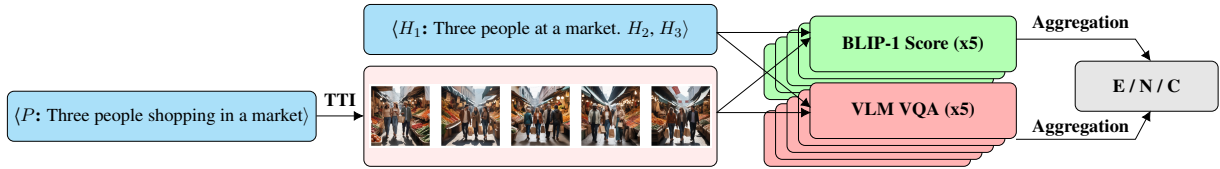}
    \caption{A diagram of the proposed method. NLI is framed in terms of the relationship between a hypothesis ($h_i$) and a visual representation of the situation depicted by a premise (p). Visual representations are obtained using a text-to-image (TTI) model. The NLI label is determined based on embedding similarity (e.g. via a model like BLIP) or by directly predicting the label in a VQA setting.}
    \label{fig:schema}
\end{figure*}

\section{Introduction}
Language models trained and fine-tuned on various textual tasks exhibit impressive performance, especially with more data. At the same time, the extent to which unimodal language models can truly represent meaning has been criticized \cite{bender-koller-2020-climbing,Bisk2020}. Indeed, while models can learn a lot about linguistic {\em form} in language, it is debatable whether exposure to text alone is sufficient to acquire {\em functional linguistic competence} — i.e., the ability to use language in real-world situations \cite{mahowald_dissociating_2024}, which also includes drawing inferences and reasoning with linguistic messages. This arguably requires models to take into account the relationship between language and the world, for instance, through visual or other perceptual channels \cite[but see][for different positions on this issue]{pavlick_symbols_2023,mandelkern_language_2024}. Enhancing language models with multimodal capabilities has now become common \cite[recent examples include][]{deitke_molmo_2024,peng_kosmos-2_2023,li_llava-onevision_2024,chen_janus-pro_2025}. The fusion of linguistic and visual modalities is often cited as a way to address the classic grounding problem \cite{harnad1990symbol}, whereby a natural or artificial agent needs to establish systematic links between symbols (say, in natural language) and elements of perception and experience. Though multimodal models perform well in several downstream tasks, it is yet to be shown whether grounding abilities can yield more robust meaning representations and reasoning abilities, obviating the need to fine-tune models on specific datasets for reasoning with natural language. 

Natural Language Inference (NLI; aka Textual Entailment) is a case in point. This task is typically framed in terms of the relationship between pairs of texts: given a premise $p$ and a hypothesis text $h$, the goal is to determine whether $h$ follows from (is entailed by) $p$, contradicts it, or whether the relationship is neutral \cite{Dagan2006,maccartney2009natural}. This task definition is highly flexible, and it has been argued that many understanding tasks lend themselves to an NLI framing \cite{white_inference_2017,poliak_collecting_2019}. Yet such a framing arguably focuses our attention on a very narrow view of inference, one that also makes models highly susceptible to learned biases that arise in text, from frequency effects to spurious correlations \cite[e.g.,][]{mccoy2019right}. An alternative view can be found in classical, truth-conditional semantic accounts, where entailment is not viewed as a relationship between texts, but as a relationship between the models or situations in which a given text holds true. Thus, if $p$ entails $h$, this is because the "world" in which $p$ holds is likely to be one in which $h$ also holds. 

In this paper, we consider whether, by exploiting the capabilities of multimodal models to render visually (some of) the possible situations in which $p$ might hold, we can obtain competitive zero-shot performance on NLI without the need for fine-tuning. In line with the observations above, we are especially interested in whether such an approach can help overcome biases that arise from the exclusively text-based treatment of NLI.

To validate our approach, we make use of data from the SNLI dataset 
  \cite[
  ][]{bowman2015large} as well as a novel, synthetic adversarial dataset.

Modern approaches to NLI, especially since the introduction of pretrained language models such as BERT \cite{devlin2019bert}, often rely on fine-tuning Transformer LMs. Despite their impressive quantitative performance, such NLI models have notable drawbacks. First, successful fine-tuning requires substantial computational resources and large datasets, which typically contain hundreds of thousands of examples. Second, fine-tuned models often perform poorly on new, unseen data due to biases and artifacts present in the training datasets \cite{nie2019adversarial,gururangan-etal-2018-annotation,mccoy2019right}. Attempts to mitigate these biases by expanding fine-tuning datasets exacerbate computational demands while still failing to fully address the issues of bias and out-of-distribution generalization. 
Finally, as noted earlier, text-based language models addressing semantic tasks like NLI have been criticized on theoretical grounds for capturing a "potentially useful, but incomplete, reflection of [...] actual meaning" \cite{bender-koller-2020-climbing}. This highlights the need for innovative approaches that leverage the strengths of existing models while reducing the computational burden of fine-tuning and simultaneously exploiting a richer notion of meaning.

Our framework, which is depicted in Figure~\ref{fig:schema} and elaborated in Section~\ref{sec:approach} below, builds on the idea that language understanding should rely on grounding language in the world, which neural models with perceptual interfaces are naturally adapted for. If a model can represent situations in which $p$ is true and assess the truth of $h$ in these situations, this should suffice for inference-making. For the sake of argument, we restrict ourselves to visual grounding with static images, the type of grounding for which pre-trained computational models are the most mature. While experimental work has long highlighted the potential of exploiting grounding for inference \cite[cf.][]{young2014image}, it is the recent progress in pre-trained grounded models that makes our approach to inference technically possible. Since our approach is zero-shot, it also does not depend on large-scale exposure to training data. Furthermore, the approach aligns with how entailment is characterized in semantic theory \cite{winter2016elements}, as a relation between the truth of the premise and the hypothesis across possible situations (represented as models).
%
%

Our study makes the following contributions:
\begin{itemize}
    \item We propose a novel method for zero-shot NLI;
    \item We show that this method attains high accuracy, comparable to that of text-only baselines;
    \item We validate the robustness of our method against surface biases inherent in fine-tuned NLI models;
    \item We release a new adversarial NLI dataset based on string overlap bias.
\end{itemize}


\section{Related work}
\label{sec:related}
\paragraph{Natural Language Inference} NLI is typically addressed as a unimodal task involving the relationship between textual premises and hypotheses. Following an early phase which considered the project from a logical perspective, for example in the FraCaS framework \cite{cooper_fracas_1996}, NLI came to be defined in probabilistic terms as a result of developments in the PASCAL RTE challenge \cite{Dagan2006}, which also gave rise to the current three-way classification between entailment, contradiction, and neutral \cite{giampiccolo_fourth_2008}. Large-scale datasets developed to address this challenge include SICK \cite{bentivogli_sick_2016}, SNLI \cite{bowman2015large}, and MultiNLI \cite{williams_broad-coverage_2018}. The present paper uses a subset of SNLI; the full dataset consists of 570k premise-hypothesis pairs, with premises sampled from an image captioning dataset (and which are hence linked to images) \cite{young2014image} and hypotheses which were crowdsourced. With the advent of large-scale pretraining, the field has witnessed a steady increase in NLI performance on standard benchmarks. For example, a fine-tuned version of DeBERTa \cite{he2020deberta} achieves around 90\% accuracy on SNLI, while RoBERTa \cite{zhuang-etal-2021-robustly} achieves 90.8\%, which goes up to 92.8\% with the addition of a self-explaining layer. More recently, large language models can be deployed in a zero-shot or few-shot fashion \cite{brown2020language},
 although their accuracy in this case remains below that of models fine-tuned on the task.
For example, on SNLI, Mistral-7B \cite{jiang_mistral_2023} has a reported accuracy of around 90\%, while SOTA performance is achieved with a few-shot version of T5 \cite{raffel_exploring_2020} further trained with synthetic data, reaching 94.7\% \cite{banerjee_first_2024}.

\paragraph{Bias and shortcut learning} Performance on NLI benchmarks, however, is subject to shortcut learning \cite{geirhos_shortcut_2020}, sensitivity to data permutations, and an inability to handle paraphrases with similar meanings \cite{schluter_when_2018}. For the specific case of SNLI, \citet{gururangan-etal-2018-annotation} found that in a significant proportion of samples, models can guess the correct label based on surface heuristics such as the presence of negation in the hypothesis. They identified a `hard' subset of SNLI where such surface heuristics are not present. In this paper, we use the hard subset in Section~\ref{sec:exp2}. Subsequent efforts have been dedicated to developing reliable evaluation methodologies to detect when models are not actually solving the inference task, with techniques ranging from stress testing \cite{naik2018stress} to adversarial examples \cite{belinkov_dont_2019,nie2019adversarial}. For example, the HANS dataset \cite{mccoy2019right} contains adversarial examples designed to ensure failure for models that rely on surface heuristics such as lexical overlap. In this paper, we develop similar adversarial examples to compare the susceptibility of unimodal and visually-grounded models to 
string similarity
heuristics (Section~\ref{sec:exp3}). 

\paragraph{Visually grounded inference} A separate line of work investigates the role of visual grounding in reasoning. This includes tasks involving reasoning about the differences between pairs of images \cite{Suhr2019,ventura2024nleyeabductivenliimages}; Visual Question Answering \cite{Antol2015,Goyal2019,hudson_gqa_2019}, where a model needs to answer a question based on an image, with some types of questions requiring models to go beyond object labeling \cite[for example, counting; e.g.][]{acharya_tallyqa_2019}; and visual commonsense reasoning \cite{Zellers2019,Park2020}. \citet{sun_unifine_2023} show that SOTA zero-shot performance on these tasks can be achieved by taking into account fine-grained visual and textual features. Conceptually closer to the original definition of the NLI task, \citet{V_SNLI} created a grounded version of SNLI by linking the premises to their original images in Flickr30k \cite{young2014image}. They found that the inclusion of visual information sometimes led to a change in the gold label for a premise-hypothesis pair, but also showed that models do not benefit significantly from the inclusion of images, relative to only using textual premises. Later approaches benefited more from grounding, but not by a large margin \cite{Kiela2019SupervisedMB,De2023TowardsIO}. In a different vein, \citet{suzuki_multimodal_2019} propose a logical formalism to represent both images and texts, in order to model entailment relationships between them. \citet{xie_visual_2019} 
presented a new visual-textual entailment task (VTE) and developed the SNLI-VE dataset, where the entailment relationship is defined purely between an image (which replaces the textual premise) and a hypothesis. As with \citet{V_SNLI}, it was observed that the grounding of image-hypothesis pairs sometimes results in a change of label compared to the original, text-only pairs in SNLI \cite{do_e-snli-ve_2021}. Lastly, \citet{reijtenbach-etal-2025-dataset} showed that generated images can be used for visual entailment just as effectively as the original images. This is similar in spirit to our approach; however, the goal of the present paper is to solve the original NLI task via a visual rendition of a {\em textual} premise, and to address whether this can help models overcome a reliance on surface heuristics. To that end, we propose two approaches that we describe in the following section.

\section{Experiments}\label{sec:approach}
We report two sets of experiments, designed to satisfy two requirements: (i) the approach requires no prior task-specific training; (ii) it is visually grounded. Our overall pipeline is depicted in Figure \ref{fig:schema}.
Given an NLI 
instance
$\langle h,p,L \rangle$, where $L$ is the label, our method consists of the following two steps.

\noindent 
{\bf 1. Visual representation}: we rely on a text-to-image generation model to create a sample of visual representations $V_p$ from premise $p$.

\noindent 
{\bf 2. Image-grounded inference}: We compare two different techniques to infer $L$ from $h$ and $V_p$:

\begin{itemize}
    \item \textbf{Cosine Similarity Score (CSS):} determines $L$ based on the similarity between embeddings of $V_p$ and $h$. We assume that high scores indicate entailment, low scores indicate contradiction, and moderate scores indicate neutrality. Given that SNLI premises are paired with at least one hypothesis of each label, $L$ can be approximated through ordering these hypotheses by similarity: the most similar $h$ is an entailment, the least similar a contradiction, and the intermediate one is neutral.

    \item \textbf{Visual Question Answering (VQA):} using a multimodal generative model, generates the most likely $L$ based jointly on $V_p$ and $h$.
\end{itemize}

Additionally, in each experiment, we attempted to estimate the impact of text-to-image bias and the extent to which it can be mitigated. We did so by generating 5 images per premise and inferring the labels from each one of those; we then aggregated the resulting labels using the following techniques: (i) \textit{Majority vote} selects the most common label, picking randomly in case of a tie; (ii) \textit{Average Value} assigns each prediction a numeric value (1 for Entailment, 0 for Neutral, -1 for Contradiction), and the 5 values are averaged; (iii) \textit{Oracle-Guided}: if any of the five predictions matches the gold label, this label is selected. Note that the oracle-guided method explores the upper performance bound of either method.

\subsection{Experiment 1}
\label{sec:exp1}

\paragraph{Data} We evaluate the proposed approach using V-SNLI, \citet{V_SNLI}'s grounded version of SNLI. For the purpose of this experiment, we select the first 100 premises and their respective hypotheses from the SNLI training split.

\paragraph{Visual Representation}
To generate visual representations $V_p$, we employ \texttt{stable-diffusion-xl-base} \cite{podell2023sdxl}. Five images are generated for each individual premise.

\paragraph{Inference} The two inference methods are implemented as follows. (i) For CSS, we use BLIP-1 \cite{BLIP1} to estimate the alignment between each premise and its respective hypotheses. (ii) For VQA, we utilize \texttt{gpt4-vision-preview}\footnote{Experiment 1 was first conducted in March and April 2024. Aside from proprietary models, we tested several alternatives: LLaVa-NeXT \cite{liu2024llavanext} and InstructBLIP \cite{dai2023instructblip}. However, we found that they struggled to consistently follow instructions given the complexity of the task, making them less suitable for our demonstration.} \cite{2023GPT4VisionSC}. However, due to the deprecation of that model, we later repeated the same experiment using \texttt{gpt-4o-2024-05-13} to ensure the reproducibility of our tests. 
We query the VQA models with an image and a textual prompt; the latter includes all three hypotheses and instructs the model to produce NLI labels for all three at once. Exposing the models to all three hypotheses ensures consistency with CSS inference, which considers the hypotheses jointly rather than independently. The full prompt is provided in Appendix \ref{sec:appendixd}. 

\paragraph{Baselines} We use three zero-shot baselines, both symbolic and neural, to test the validity of visual grounding for NLI. Much like the CSS approach, our baselines rank hypotheses based on their similarity to $p$ and infer the labels accordingly: entailment is the most similar, contradiction is the least similar, and neutral is in between. However, they only consider textual features and differ in the feature extraction method:

\noindent
\textbf{BLEU}: We compute BLEU~\cite{Papineni2002,post-2018-call} between $p$ and $h$ and consider it as a measure of similarity between them. Good performance based on BLEU would suggest that n-gram overlap heuristics contribute to solving the NLI task. 

\noindent
\textbf{NSP}: This baseline takes advantage of the next sentence prediction capabilities of BERT-like models. \citealt{devlin2019bert} demonstrate that NSP pre-training enhances the performance of BERT on NLI. In our experiment, hypotheses are ranked based on how probable they are as the next sentence after the premise, according to the BERT NSP classifier.

\noindent 
\textbf{BERTcss}: We extract {\tt CLS} embeddings for both $p$ and $h$ using BERT-base and compute the cosine similarity between them to rank hypotheses.

\begin{table}[!t]
  \small
  \centering
  \begin{tabular}{|l|l|ll|}
    \hline
    \textbf{Task} & \textbf{Method} & \textbf{Aggregation}& \textbf{Score} \\
    \hline
    \multirow{9}{*}{NLI}  & \multirow{3}{*}{CSS} & Oracle & 79.1\% \\ 
     &  & Average Val. & 69.0\% \\
     &  & Maj. Class & 69.4\% \\\cline{2-4}
     & \multirow{3}{*}{VQA$_{GPT4V}$} & Oracle & 81.0\% \\
     &  & Average Val. & 77.0\% \\
     &  & Maj. Class & 74.3\% \\\cline{2-4}
     & \multirow{3}{*}{VQA$_{GPT4o}$} & Oracle & 80.0\% \\
     &  & Average Val. & 74.3\% \\
     &  & Maj. Class & 73.0\% \\
    \hline
    \hline
    \multirow{10}{*}{Ent.}  & \multirow{3}{*}{CSS} & Oracle & 78.6\% \\  
     &  & Average Val. & 68.7\% \\
     &  & Maj. Class & 66.9\% \\\cline{2-4}
     & \multirow{3}{*}{VQA$_{GPT4V}$} & Oracle & 95.1\% \\
     &  & Average Val. & 87.3\% \\
     &  & Maj. Class & 90.2\% \\\cline{2-4}
     & \multirow{3}{*}{VQA$_{GPT4o}$} & Oracle & 92.0\% \\
     &  & Average Val. & 81.1\% \\
     &  & Maj. Class & 85.1\% \\
    \hline
    \hline
    \multirow{10}{*}{Contr.}  & \multirow{3}{*}{CSS} & Oracle & 87.9\% \\ 
     &  & Average Val. & 82.6\% \\
     &  & Maj. Class & 82.8\% \\\cline{2-4}
     & \multirow{3}{*}{VQA$_{GPT4V}$} & Oracle & 98.9\% \\
     &  & Average Val. & 92.9\% \\
     &  & Maj. Class & 94.9\% \\\cline{2-4}
     & \multirow{3}{*}{VQA$_{GPT4o}$} & Oracle & 98.0\% \\
     &  & Average Val. & 94.0\% \\
     &  & Maj. Class & 96.0\% \\
    \hline
    \hline
    \multirow{10}{*}{Neut.}  & \multirow{3}{*}{CSS} & Oracle & 70.5\% \\  
     &  & Average Val. & 56.0\% \\
     &  & Maj. Class & 57.9\% \\\cline{2-4}
     & \multirow{3}{*}{VQA$_{GPT4V}$} & Oracle & 47.9\% \\
     &  & Average Val. & 50.0\% \\
     &  & Maj. Class & 36.7\% \\\cline{2-4}
     & \multirow{3}{*}{VQA$_{GPT4o}$} & Oracle & 49.5\% \\
     &  & Average Val. & 47.5\% \\
     &  & Maj. Class & 37.4\% \\
    \hline
  \end{tabular}
  \caption{\label{tableEx1}
  Accuracy scores for experiment 1 (\textbf{5 images} per premise) overall (NLI) and per-class.}
\end{table}

\begin{table*}[!t]
  \centering
  \small
  \begin{tabular}{|l|lll|lll|}
    \hline
    \textbf{Task} & CSS& VQA-1 & VQA-2 & BLEU & NSP & BERTcss \\
    \hline
    NLI  & 69.0\% & 74.6\% & 74.3\% & 39.0\% & 47.0\% & 42.0\% \\
    Entailment & 69.7\% & 90.1\% & 86.1\% & 44.9\% & 50.0\% & 47.9\% \\
    Contradiction & 80.6\% & 95.0\% & 97.9\% & 43.8\% & 56.2\% & 41.0\% \\
    Neutral & 57.0\% & 37.7\% & 39.3\% & 28.8\% & 36.9\% & 38.2\% \\

    \hline
  \end{tabular}
  \caption{\label{table_single_image}
    Performance of CSS and VQA against baselines on 100 premises (\textbf{1 image} per premise), overall (NLI) and per-class.}
\end{table*}

\subsubsection{Results}\label{results_experiment1}

Results of Experiment 1 are summarized in Tables~\ref{tableEx1} \&~\ref{table_single_image}.  
They demonstrate that image-based methods can attain meaningful classification accuracy. Furthermore, they show that visual information is primarily effective at distinguishing entailment and contradiction. It is also evident that image-based inference struggles to handle neutral instances well, resulting in a sizeable accuracy drop. Aggregating predictions based on multiple (diverse) images leads to better accuracy on that class (compare Tables ~\ref{tableEx1} \& ~\ref{table_single_image}).

\subsection{Experiment 2: Biased Data}\label{sec:exp2}
An important argument in favor of the method we propose is that we expect it to be robust against surface heuristics compared to task-specific fine-tuning.
Nevertheless, one of our individual pipelines, VQA, could potentially be affected by annotation artifacts in hypotheses and use NLI-relevant heuristics, such as negation, to shortcut on image analysis. This could have impacted the results of the first experiment, in which we did not account for possible biases of this kind. The second experiment is designed to address the question of hypothesis-level bias.

\paragraph{Data} In experiment 2, we draw on \citet{gururangan-etal-2018-annotation}'s distinction between "hard" and "easy" subsets of SNLI data (see Section \ref{sec:related}). 
We sample 300 hypotheses related to 100 premises from both subsets. When more than three hypotheses were associated with one premise, we discarded the surplus hypotheses. This resulted in 276 "easy" hypotheses related to 92 premises and 285 "hard" hypotheses related to 95 premises. In what follows, we refer to these as the "easy" and "hard" subsets.

\paragraph{Visual Representation}
To generate the images $V_p$ from premises, we made use of two alternative text-to-image (TTI) models: Stable Diffusion and DALL-E 3 \cite{BetkerImprovingIG}, a state-of-the-art image generation model at the time these experiments were conducted (September 2024). Our intuition was that leveraging DALL-E images would improve inference accuracy, since this model is less prone to concept bleeding\footnote{Concept bleeding occurs when, given a predicate that applies to a specific argument, a text-to-image model renders the image such that the property denoted by the predicate applies to other arguments as well \cite{podell2023sdxl}.} and other potentially misleading TTI generation errors (see Section~\ref{sec:err}). As in the first experiment, we additionally considered the results of aggregated inference based on 5 images per premise.

\paragraph{Inference} The inference procedure largely followed the first experiment for both the CSS and the VQA approaches. Since \texttt{gpt4-vision-preview} was no longer available, VQA experiments were conducted with \texttt{gpt-4o-2024-05-13}.

\paragraph{Baseline} In the second experiment, our objective was to determine whether multimodal or unimodal inference methods handle bias more efficiently. Our main text-only baseline is NLI fine-tuned RoBERTa \cite{zhuang-etal-2021-robustly}, a part of the \texttt{sentence-transformers} library \cite{reimers-2019-sentence-bert}. RoBERTa was trained and fine-tuned on a known set of data, and it is certain that it has not been exposed to the test split of SNLI. At the same time, RoBERTa performs comparably to state-of-the-art models on the NLI task. 
As an additional sanity check baseline, we once again use BLEU, which only considers surface overlap.

\begin{table}[!t]
  \centering
  \small
  \begin{tabular}{|l|lll|}
    \hline
    \textbf{Task} & \textbf{Method} & \textbf{Easy} &  \textbf{Hard} \\
    \hline
    \multirow{6}{*}{NLI}  & BLEU & 39.1\% & 44.6\%{ \small (+5.5)} \\ 
     & RoBERTa & 98.9\%  & 83.1\%{ \small (-15.8)} \\ 
     & GPT-4O & 96.3\% & 85.6\%{ \small (-10.7)} \\
     & CSS-DALL-E & 70.3\% & 69.1\%{ \small (-1.2)} \\
     & CSS-SD & 71.7\% & 65.2\%{ \small (-6.5)} \\
     & VQA-DALL-E & 90.6\% & 74.7\%{ \small (-15.9)} \\
     & VQA-SD & 89.5\% & 70.2\%{ \small (-19.3)} \\
    \hline
    \hline
    \multirow{6}{*}{\small Ent.} & BLEU & 41.0\% & 48.2\%{ \small (+7.2)} \\
     & RoBERTa & 98.9\% & 81.6\%{ \small (-17.3)} \\ 
     & GPT-4O & 95.8\% & 86.8\%{ \small (-9)} \\
     & CSS-DALL-E & 64.2\% & 63.1\%{ \small (-1.1)} \\
     & CSS-SD & 65.2\% & 57.0\%{ \small (-8.2)} \\
     & VQA-DALL-E & 93.7\% & 78.1\%{ \small (-15.6)} \\
     & VQA-SD & 87.4\% & 73.7\%{ \small (-13.7)} \\
    \hline
    \hline
    \multirow{6}{*}{\small Contr.} & BLEU & 46.2\% & 52.5\%{ \small (+6.3)} \\ 
     & RoBERTa & 97.8\% & 86.1\%{ \small (-11.7)} \\ 
     & GPT-4O & 98.9\% & 92.1\%{ \small (-6.8)} \\
     & CSS-DALL-E & 87.1\% & 82.2\%{ \small (-4.9)} \\
     & CSS-SD & 89.2\% & 82.2\%{ \small (-7)} \\
     & VQA-DALL-E & 97.8\% & 93.1\%{ \small (-4.7)} \\
     & VQA-SD & 97.8\% &  91.1\%{ \small (-6.7)} \\
    \hline
    \hline
    \multirow{6}{*}{\small Neut.} & BLEU & 29.5\% & 27.1\%{ \small (-2.4)} \\ 
     & RoBERTa & 100.0\% & 81.4\%{ \small (-18.6)} \\ 
     & GPT-4O & 94.3\% & 74.3\%{ \small (-20)} \\
     & CSS-DALL-E & 59.1\% & 60.0\%{ \small (+0.9)} \\
     & CSS-SD & 60.2\% & 54.2\%{ \small (-6)} \\
     & VQA-DALL-E & 79.5\% & 42.8\%{ \small (-36.7)} \\
     & VQA-SD & 82.9\% &  34.3\%{ \small (-48.6)} \\
    \hline
  \end{tabular}
  \caption{\label{tableEx2} Overall and per-class percentage accuracy for experiment 2 on SNLI easy and hard subsets (\textbf{1 image} per premise). In parentheses: change in accuracy between easy and hard subsets.}
\end{table}

\subsubsection{Results}
We report the results in Table \ref{tableEx2}. One trend that we observe is that both RoBERTa and the VQA model show considerably lower accuracy on the hard compared to the easy subset, suggesting that both models might rely on surface heuristics related to the hypothesis text $h$. In contrast, CSS shows a much smaller gap in performance between the two subsets; however, it also demonstrates lower overall accuracy. Regarding specific classes, distinguishing contradictions remains relatively easy for both VQA and CSS (to the extent that VQA surpasses RoBERTa on the "hard" set), but the Neutral class continues to be challenging to handle. 

The results also reveal another tendency: the BLEU baseline is \textit{more} accurate on the hard subset, compared to the easy one. This suggests that while the hard subset mitigates the utility of heuristic biases on the hypothesis text, it may be more susceptible to another heuristic, namely the lexical overlap between $p$ and $h$, an issue we return to in Section~\ref{sec:exp3}.

\begin{table}[htp]
  \small
  \centering
  \begin{tabular}{|l|l|ll|}
    \hline
    \textbf{Task} & \textbf{Method} & \textbf{Aggregation}& \textbf{Score} \\
    \hline
    \multirow{6}{*}{NLI}  & \multirow{3}{*}{VQA-DALL-E} & Oracle & 83.1\% \\
     &  & Average Val. & 74.4\% \\
     &  & Maj. Class & 74.4\% \\\cline{2-4}
     & \multirow{3}{*}{VQA-SD} & Oracle & 78.2\% \\
     &  & Average Val. & 70.2\% \\
     &  & Maj. Class & 71.6\% \\
    \hline
    \hline
    \multirow{6}{*}{Ent.}  & \multirow{3}{*}{VQA-DALL-E} & Oracle & 89.5\% \\
     &  & Average Val. & 76.3\% \\
     &  & Maj. Class & 79.8\% \\\cline{2-4}
     & \multirow{3}{*}{VQA-SD} & Oracle & 85.1\% \\
     &  & Average Val. & 71.0\% \\
     &  & Maj. Class & 77.2\% \\
    \hline
    \hline
    \multirow{6}{*}{Contr.}  & \multirow{3}{*}{VQA-DALL-E} & Oracle & 96.0\% \\
     &  & Average Val. & 93.0\% \\
     &  & Maj. Class & 93.0\% \\\cline{2-4}
     & \multirow{3}{*}{VQA-SD} & Oracle & 93.0\% \\
     &  & Average Val. & 91.0\% \\
     &  & Maj. Class & 91.0\% \\
    \hline
    \hline
    \multirow{6}{*}{Neut.}  & \multirow{3}{*}{VQA-DALL-E} & Oracle & 54.3\% \\
     &  & Average Val. & 44.3\% \\
     &  & Maj. Class & 38.6\% \\\cline{2-4}
     & \multirow{3}{*}{VQA-SD} & Oracle & 45.7\% \\
     &  & Average Val. & 38.6\% \\
     &  & Maj. Class & 34.3\% \\
    \hline
  \end{tabular}
  \caption{\label{tableEx6} Overall and per-class percentage accuracy on SNLI-hard in Experiment 2. (\textbf{5 images} per premise)}
\end{table}

Some evidence for the impact of visual grounding comes from the use of multiple images. The results of aggregation over several images, provided in Table \ref{tableEx6}, show that averaging and oracle aggregation yield more accurate predictions compared to those based on a single image. However, we also observe that TTI does not consistently yield diverse images for a given premise,
which limits the effectiveness of aggregation; we return to this point in our error analysis (Section~\ref{sec:err}). Additionally, we note the influence of the text-to-image generation procedure: in the VQA setting, generating images with DALL-E leads to noticeably better accuracy compared to Stable Diffusion.

\subsubsection{Quantifying hypothesis-side bias} Data from SNLI-easy and SNLI-hard can help us quantify how much different models depend on hypothesis-level heuristics, disregarding the premise. This question is not fully answered by comparing the performances on the two subsets, because models may still rely on information in an informative premise, even when the hypothesis alone can be leveraged to solve the task.

To mitigate this, we conduct a separate experiment where we replace all premises with an uninformative statement, "Something is happening." A model that can correctly predict the original $L$ from this altered $p$ and the original $h$ must be exploiting hypothesis-side heuristics. These are known to be present in the easy subset and absent in the hard subset. Therefore, with $p$-s replaced, the delta between a model's accuracy on "easy" vs.\ "hard" accurately reflects how often our models make predictions based on hypotheses alone.

We compare VQA-DALL-E, CSS-DALL-E, and RoBERTa based on these criteria.
Since the predictions of VQA-DALL-E and CSS-DALL-E can vary substantially depending on the image, we report the average accuracy derived from five images. 

As shown in Table \ref{table_something}, RoBERTa exhibits a wide delta of 23.3\% and performs below random on hard hypotheses, indicating a substantial hypothesis-side bias. 
In contrast, VQA-DALL-E and CSS-DALL-E demonstrate narrower deltas than the text-based model:
8.5\% and 12.2\%, respectively. VQA-DALL-E's and CSS-DALL-E's accuracy on hard hypotheses remains close to random, indicating some reliance on hypothesis properties but no strong bias.

Qualitatively, both methods tend to classify most examples as contradictions. Manual inspection suggests that VQA pipeline only relies on the heuristic involving the mention of thoughts, intentions, and other non-visual predicates in the hypothesis (e.g., "Boys \textbf{trying to escape} an incoming storm"), which leads it to classify such examples as neutral. This behavior may be influenced by the prompt, where this type of predicates is explicitly mentioned as a criterion for the Neutral class: see Appendix \ref{sec:appendixd}. RoBERTa, on the other hand, appears to consider additional properties of the hypothesis, such as the occurrence of modifiers.

\begin{table}
  \centering
  \small
  \begin{tabular}{|l|ll|}
    \hline
    \textbf{Method} & \textbf{Easy}  & \textbf{Hard} \\ 
    \hline
    Random & 33.33\% & 33.33\% (0\%) \\
    \hline
    RoBERTa & 51.4\% & 28.1\%  (-23.3\%)\\ 
    CSS-DALL-E & 42.0\% & 29.8\%  (-12.2\%)\\ 
    VQA-DALL-E & 45.5\% & 37.0\% (\textbf{-8.5}\%) \\
    \hline
  \end{tabular}
  \caption{\label{table_something}
    Accuracy on uninformative premises (averaged over \textbf{5 images} for VQA); deltas (in parentheses) estimate the degree of hypothesis side bias.}
\end{table}

\subsection{Experiment 3: Adversarial Data}\label{sec:exp3}
Experiment 2 revealed a potential source of bias in the hard SNLI subset, beyond the hypothesis-side heuristics. The higher accuracy of the BLEU baseline on the hard subset suggests the presence of considerable textual overlap between $p$ and $h$, which could provide an additional shortcut for language models to exploit.
Our third experiment was designed to address this question and investigate whether our methods are susceptible to this bias.


\paragraph{Data} Inspired by previous work \cite[e.g., 
][]{mccoy2019right}, we designed a synthetic dataset intended to mitigate overlap-based heuristics, while ensuring that premises could be rendered accurately as images. By creating a controlled synthetic dataset, we aimed to achieve two goals: first, address both lexical and sequence overlap as NLI heuristics; second, analyze on a small scale how frequently incorrect text-to-image generation leads to errors.

Sample premises were generated according to the following template: {\em The \texttt{[noun1]} who \texttt{[transitive\_verb]} the \texttt{[noun2]} \texttt{[intransitive\_verb]}.} (e.g., {\em The girl who greets the dog laughs.}). From such premises, an entailed hypothesis was generated based on the template {\em The \texttt{[noun1]} \texttt{[intransitive\_verb]}.} (e.g., {\em The girl laughs.}); similarly, non-entailment examples were generated based on the template {\em The \texttt{[noun2]} \texttt{[intransitive\_verb]}} (e.g., {\em The dog laughs.}). In order to reduce ambiguity, we did not generate neutral statements. We expected the hypotheses to be adversarial concerning both lexical overlap and subsequence, since both of them share an equal number of words with the premise, but \textit{\texttt{[noun2]}} also directly precedes \textit{\texttt{[intransitive\_verb]}}, prompting a heuristic-influenced model to likely recognize it as the verb's subject.
\noindent
Nouns and verbs were chosen from a small, manually selected set that maximized the chances of accurate visual representation: \textit{\texttt{[intransitive\_verb]}} values denoted actions with clear facial expressions (e.g., {\em laughs}); the two noun values consisted of a human- and an animal-denoting noun. Initial experiments with two human-denoting nouns with distinct visual features (e.g., {\em policeman} and {\em mechanic}) showed that Stable Diffusion and, to a lesser extent, DALL-E were prone to introducing concept bleeding, whereby both characters would share the same facial expression described by \textit{\texttt{[intransitive\_verb]}}.
In total, we generated 100 premises, each paired with both an entailed and a non-entailed hypothesis, resulting in a total of 200 pairs\footnote{We will make our data available on GitHub after the anonymity period.}. 

\paragraph{Baseline} Once again, we include fine-tuned RoBERTa as a strong comparison. The BLEU-based baseline used in experiment 2 is not applicable here, as the lexical overlap between hypotheses and premises is identical for both entailments and non-entailments.

\paragraph{Inference} The pipelines for the compared models remained the same as in Experiment 2, utilizing DALL-E-3, BLIP-1, and GPT4o. The only adjustment was that  both models operated with a set of two labels ("entailment" and "non-entailment") instead of the three standard NLI labels. The VQA prompt was updated accordingly.

\begin{table}
  \centering
  \begin{tabular}{|l|ll|}
    \hline
    \textbf{Task} & \textbf{Method}  & \textbf{Adversarial} \\ 
    \hline
    {\small Ent. vs Non-Ent.} & RoBERTa & 65.5\% \\ 
     & CSS-DALL-E & 56.0\% \\
     & VQA-DALL-E & 85.0\% \\
     & VQA-SD & 74.0\% \\
    \hline
  \end{tabular}
  \caption{\label{table_adversarial}
    Percentage accuracy on adversarial data, Experiment 3 (\textbf{1 image} per premise).}
\end{table}

\subsubsection{Results}
VQA achieves an accuracy of 85\% on the adversarial data using DALL-E images, outperforming the fine-tuned RoBERTa model (65.5\%), see Table~\ref{table_adversarial}.
In contrast, CSS yields a lower accuracy of 56\%, indicating a noticeable word overlap bias. A possible explanation for this is that BLIP-1 focuses on matching tokens with image regions while losing syntactic information. We discuss this issue in more detail in Section \ref{sec:err}.
We conducted a manual analysis of the generated images whereby we found that up to 15\% of these suffered from concept bleeding. This issue largely overlaps with the erroneous predictions made by VQA: out of 30 errors, 14 were due to factual inaccuracies in the images, 11 due to concept bleeding, and only 5 due to incorrect image-to-text inference. These observations further highlight that improvements in TTI generation will enhance the performance of our method.

\section{Error analysis}\label{sec:err}
We distinguish two general types of errors in our pipeline: those that stem from visualizing NLI premises (1) and those from the subsequent inference (2).

\subsection{Text-to-image: Factual Errors}
Text-to-image models have some widely known limitations, such as factual inaccuracies, concept bleeding, and incorrect object counts \cite{podell2023sdxl}, which we also encountered in practice for some of the SNLI examples. When this occured, the inference often led to incorrect results. One such example is illustrated in Figure \ref{fig:ex1}: "One man sits inside and plays the banjo; there are trees behind him outside" is depicted as an outdoor scene. Despite this, the fact that DALL-E is less prone to such issues compared to SD-XL suggests that more advanced text-to-image models may become increasingly robust in this respect.

\subsection{Text-to-image: Neutrality Errors}
When constructing hypotheses for the neutral class, SNLI annotators employed a limited set of strategies \cite{gururangan-etal-2018-annotation}. One common strategy observed in our subsets of SNLI involves introducing an object that is semantically fitting but not mentioned in the premise. For instance, consider the premise "Three men standing on grass \textbf{by the water} looking at something on a table" and the hypothesis "The three men are \textbf{by the lake}". While the textual premise can remain agnostic about specific details, such as the type of water, translating this premise into the visual domain forces the model to depict a specific landscape, such as a seashore, a lake-shore or a riverbank (as illustrated by Figure \ref{fig:ex2}). Consequently, neither of our methods can reliably identify the neutral class. The same issue affects many other neutral instances as well.

The multi-image approach described in Section \ref{sec:approach} could potentially address this issue: assuming that the problematic object appears in some images but not in others, aggregating the inference results could lead to the neutral class being selected as a middle ground. 
However, in our evaluation attempts, even with varying random seeds and temperature values, the model-generated images either consistently omitted or consistently included the problematic visual objects ("lake", etc.). As a result, normalizing predictions over sets of five images, whether through averaging or majority voting, did not yield the expected improvement in scores.

\subsection{CSS: Object overlap}
\begin{figure}
    \centering
    \begin{subfigure}{1\linewidth}
    \includegraphics[width=1\linewidth]{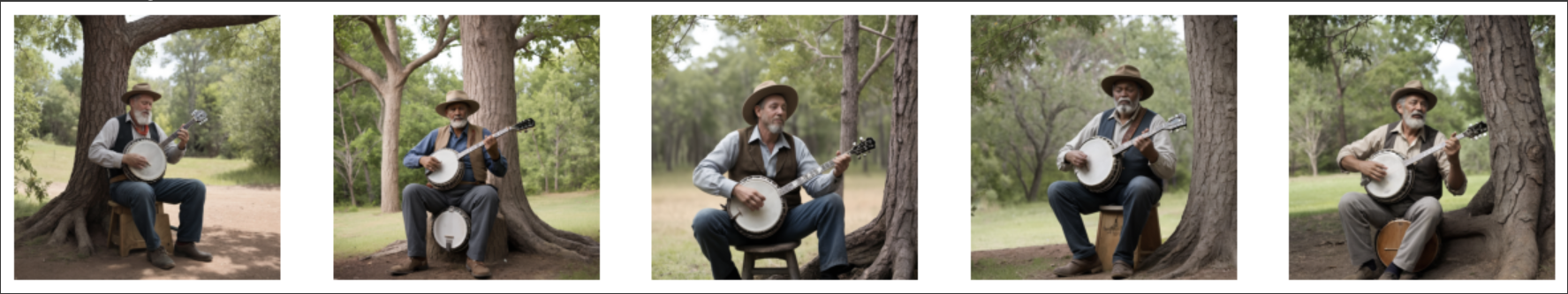}
    \caption{}
    \label{fig:ex1}
    \end{subfigure}
    \begin{subfigure}{1\linewidth}
    \includegraphics[width=1\linewidth]{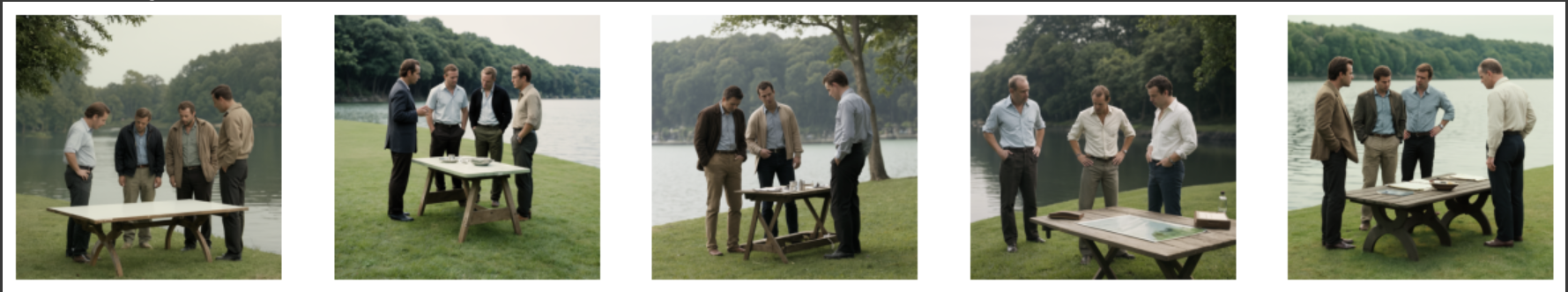}
    \caption{}
    \label{fig:ex2}
    \end{subfigure}
    \begin{subfigure}{1\linewidth}
    \includegraphics[width=1\linewidth]{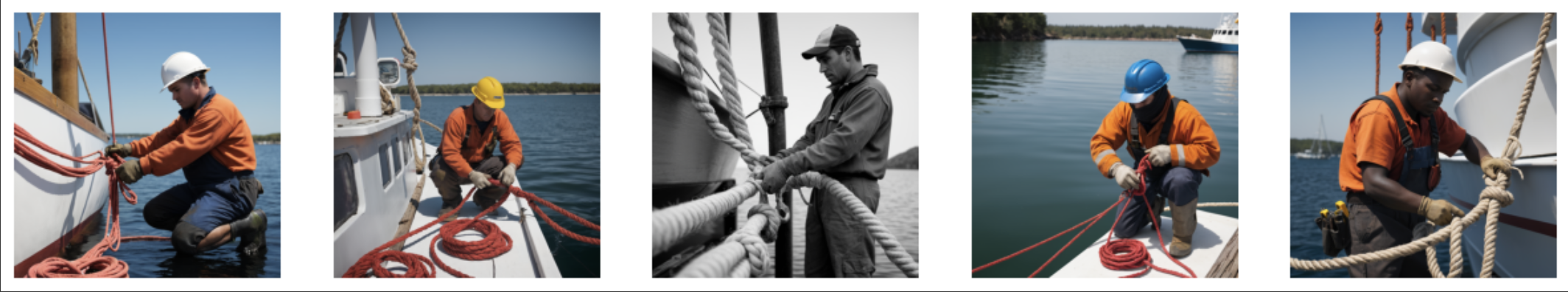}
    \caption{}
    \label{fig:ex3}
    \end{subfigure}
    \caption{\label{error_examples}
    Examples of incorrectly classified premise representations.}
\end{figure}

Powered by BLIP, CSS is prone to a multimodal overlap bias: it assigns higher similarity scores to descriptions that include more words with visual referents in the image, regardless of their semantic correctness. \citealp{pezzelle2023dealingsemanticunderspecificationmultimodal} observes a similar tendency for CLIP. Likewise, models of this kind are also unable to fully leverage compositional information due to how they handle text and image encoding~\cite{kamath_text_2023}. An example of this can be found in Figure \ref{fig:ex3}, where for $p$ "A boat worker securing a line", CSS favors the neutral $h$, "The boat worker works hard to establish the line", and not the entailment, "A worker is doing something to a boat". The "line" object, present in the neutral hypothesis and in the images, contributes to a higher overlap.

\section{Conclusion}
\label{sec:conclusion}
In this paper, we present the first implementation of an NLI classifier based purely on visual grounding. Our method is zero-shot and exhibits high accuracy, surpassing the few-shot results reported for NLI in \citet{brown2020language}.  Unlike state-of-the-art task-specific fine-tuning, our methods (VQA and, to a lesser extent, CSS) are cognitively and theoretically grounded and avoid known superficial biases. They are also robust against injecting irrelevant biases from training data (although the specific implementation of CSS tested here exhibits overlap effects). While the standard approach of fine-tuning models on adversarial data can mitigate biases (and potentially introduce new ones), we demonstrated how these biases can be mitigated by \emph{avoiding} fine-tuning. Due to the use of images as an intermediate representation, our method offers additional interpretability. For example, in our error analysis, we were able to track errors down to specific components of our system, such as text-to-image generation or image-based inference.

We tested our method on examples describing everyday scenes as represented by SNLI. With necessary modifications, our approach can be extended to other reasoning tasks where relevant information can be visualized, such as spacial reasoning. At the same time, as is the case with visual grounding per se, dealing with abstract entities remains challenging \cite{beinborn-etal-2018-multimodal}.

A notable limitation of our approach is that it works much better with entailments and contradictions than with neutral hypotheses. We believe that aggregation of predictions from multiple diverse images can help bridge this gap; our preliminary results in this directions are cautiously promising.

Many of the errors we observed are linked to limitations in the existing image generation models. Therefore, one can expect that the same basic method will demonstrate significant improvements over time, given the rapid pace at which image generation models are evolving. Furthermore, we find it promising to eventually extend this approach to
more diverse types of grounding, such as audio or video.

We argued in this paper that grounding can provide a radically different approach to NLI. However, our general hypothesis suggests that grounding offers a qualitatively better handling of meaning in general. Can the essence of our method extend to other tasks? One could envision, for instance, story generation with interleaved generated illustrations that also help support the narrative coherence, or visualization as an aid for understanding and generating metaphorical language.

In the context of NLI, our approach reduces training compute but adds more steps at inference time, including image generation. Interestingly, the current practice of text-only models also trends toward increased test-time compute, sometimes to a much greater extent than our approach \cite{openai-o1}. 

Lastly, our approach might have further theoretically interesting properties.
First, while we tested our method on predicting categorical NLI labels,  relying on multiple images and possibly on the continuous nature of image-hypothesis match can address the probabilistic nature of NLI, recognized since the first competitions on the task \cite{glickman2005web}.
Second, our work can contribute to old theoretical debates. Take the example of two views on semantics \cite{lepore1983model}, model theoretic (based on reference) vs.\ structural (based on formal relations between expressions that support semantic inference). Grounded modeling of meaning like in our method corresponds to model theoretic semantics while language modeling based approaches correspond to structural semantics. The two might therefore not only be theoretically complementary but also computationally implemented differently using state-of-the-art AI models.

\section*{Acknowledgments}

\bibliography{latex/custom.2}
\bibliographystyle{acl_natbib}

\appendix

\section{Appendix VQA Prompt}\label{sec:appendixd}
{\em In line with recommendations by \citet{sun-etal-2024-exploring}, we replace the task-specific labels (entailment/contradiction/neutral) with natural language labels (accurate/contradicting/neither). }

Question: With respect to the objects in the image, is the statement in square brackets a) accurate, b) contradicting, c) neither?

Answer 'accurate' if the statement accurately describes the objects in the image.

Answer 'contradicting' if objects miss from the image or if the description is incorrect.

Answer 'neither' if the statement is factually correct, but makes claims that don't follow from the image (intentions, social relations etc.). 

Put your answers into angle brackets.

Statement 1: [...]

Statement 2: [...]

Statement 3: [...]

Answer 1 (accurate/contradicting/neither): <...>

Answer 2 (accurate/contradicting/neither): <...>

Answer 3 (accurate/contradicting/neither): <...>

\end{document}